\documentclass[twocolumn]{article}
\usepackage[top=2cm, bottom=2.5cm, left=2cm, right=2cm]{geometry}
\usepackage{times,textcomp}
\usepackage[T1]{fontenc}
\usepackage[utf8]{inputenc}
\usepackage{amssymb,amsmath}
\usepackage{enumitem}
\usepackage{eurosym}
\usepackage{listings}
\usepackage{float,caption}
\usepackage{diagbox}
\usepackage{tabularx}

\usepackage[giveninits=true]{biblatex}
\setlist[itemize]{itemsep=0pt}

\usepackage{titlesec} 
\titleformat{\section}{\large\bfseries}{\thesection}{1em}{\MakeUppercase} 
\titleformat{\subsection}{\large\bfseries}{\thesubsection}{1em}{}
\titlespacing*{\section}{0pt}{\parskip}{0pt}
\titlespacing*{\subsection}{0pt}{\parskip}{0pt}

\usepackage{longtable,booktabs}
\usepackage{graphicx,grffile}
\usepackage{fancyhdr}

\title{Evaluating categorical encoding methods on a real credit card fraud detection database}

\usepackage{authblk}

\author{François de la Bourdonnaye, Fabrice Daniel}

\affil{\small Artificial Intelligence Department of Lusis, Paris, France\\http://www.lusisai.com}

\date{December 2021}
\begin{document}

\maketitle

\begin{abstract}
Correctly dealing with categorical data in a supervised learning context is still a major issue. Furthermore, though some machine learning methods embody builtin methods to deal with categorical features, it is unclear whether they bring some improvements and how do they compare with usual categorical encoding methods.
In this paper, we describe several well-known categorical encoding methods that are based on target statistics and weight of evidence. We apply them on a large and real credit card fraud detection database. Then, we train the encoded databases using state-of-the-art gradient boosting methods and evaluate their performances. We show that categorical encoding methods generally bring substantial improvements with respect to the absence of encoding. The contribution of this work is twofold: (1) we compare many state-of-the-art "lite" categorical encoding methods on a large scale database and (2) we use a real credit card fraud detection database.
\end{abstract}


\section{Introduction}

Credit card frauds happen everyday and represent a major cost for banks. Consequently, it becomes beneficial for them to be able to automatically detect a large amount of these frauds and block them before having to reimburse clients. A good hint for this is the use of supervised learning techniques which have known major progresses over the past years \cite{lucas2020credit}, \cite{10.1007/978-3-319-71249-9_2}, \cite{articlepozml}. Despite these hopes, several issues related to the concept drift, explainability or imbalanced dataset are still unsatisfactory addressed. In this paper, we are interested in the very specific problem of encoding categorical variables for fraud detection problems. Indeed, lot of input features of fraud detection problems are categorical-typed, e.g. the merchant category, the country in which the transaction takes place, the type of card, ...

Handling categorical data is not an easy task because we cannot naively input categorical data in machine learning pipelines without processing and hope that black boxes will do the trick in an optimal way. Several ways exist among which using target statistics, using polynomial coding, sum coding, one-hot encoding... In this paper, we mainly focus on methods using target statistics and weight of evidence because they transform categorical variables into a single numerical variable, which satisfies some size constraints: the fraud detection real dataset is huge, so the memory consumption of the encoded features should be as small as possible. In addition an important categorical feature like MCC (Merchant Category Code) has hundreds of unique values, making one-hot encoding irrelevant not only for a question of memory consumption but also for the resulting sparcity. The machine learning methods used in this study are based on gradient boosting. Three baseline methods are used: LightGBM, CatBoost and XGBoost. The choice of these techniques might be surprising given the fact literature often states that algorithms based on decision trees do not need a specific encoding for categories because of the "split" aspect, i.e. specific categories can be discriminated. We claim that this assumption may be inaccurate and will conduct experiments on real data to empirically show that models based on decision trees benefit from categorical encoding methods.

The studied encoding methods will be presented and compared against each other to check which method yields best performances. Our work contributes in that such categorical encoding methods have not been compared against each other on a huge real fraud detection database to our knowledge. Furthermore, these methods will be tested also against builtin methods when they exist so that we can conclude whether encoding categorical data with external methods is superior to builtin encoding. To finish with, the encoding methods are tested against the absence of encodings to ensure that metrics are better using categorical encoding and that our assumption of the usefulness of categorical encodings for gradient boosting trees is correct.

The remainder of the paper is organized as follows. First, we present theoretical knowledge that includes a presentation of gradient boosting, categorical encoding methods using target statistics, weight of evidence and builtin support for gradient-boosting based methods. Second, we present the experiments that are conducted. And finally, we will conclude the paper.

\section{Background}
\subsection{Gradient-boosting}

\subsubsection{Main idea}
Boosting algorithms combine weak learners (generally decision trees) in an iterative fashion. Gradient boosting algorithms have a specific scheme where a new learner is optimized such that it fits residual of the previous learner. 

Let's describe it in mathematical terms. We use the following notations:
\begin{itemize}
    \item $F_{i} (i \in \{1,...,m\})$ is a learner
    \item $D_{\mathrm{train}} = \{(x_{1},y_{1}),...,(x_{i},y_{i}),...,(x_{n_{train}},y_{n_{train}})\} = (X,Y)$ a training dataset.
\end{itemize}

At the first iteration, $F_{1}$ is fitted as a normal decision tree, i.e. on $y$. For the following iterations $i \in \{2,...,m\}$, $F_{i}$ is fitted on $y - F_{i-1}$. Thus, the rationale is to fit estimators on the residual error of the previous estimator. 

The name "gradient boosting" comes from the fact the residual is (up to a factor) the gradient of the mean squared error between the target and the output of the learner (see Equation \ref{eqgb} and \ref{eqdgb}).

\begin{equation}
L_{\mathrm{MSE}} = \frac{1}{n_{train}}(y-F_{m}(x))^{2}
 \label{eqgb}  
\end{equation}

\begin{equation}
\frac{\delta L_{\mathrm{MSE}}}{\delta F_{m}(x)} = \frac{2}{n_{train}}(y-F_{m}(x))
 \label{eqdgb}  
\end{equation}

\subsubsection{XGBoost}
XGBoost \cite{Chen_2016} is one of the most famous gradient boosting libraries. Its keys innovations are based on a novel tree learning algorithm that efficiently handles sparse data, "a theoretically justified weighted quantile sketch procedure that enables handling instance weights in approximate tree learning", the efficient use of ressources for parallel computing, and "an  effective  cache-aware  block  structure or out-of-core tree learning". 

\subsubsection{LightGBM}

The major difficulty of most gradient boosting algorithms is their scalability when the number of features increases. LightGBM \cite{10.5555/3294996.3295074} exhibits two techniques that are derived to overcome this difficulty: 
\begin{itemize}
    \item Gradient-based one side sampling (GOSS): this procedure cancels data that have no gradients.
    \item Exclusive feature bundling: this method bundles mutually exclusive features, i.e. features that never take 0 simultaneously.
\end{itemize}
Note that we won't use the GOSS feature for our experiments.
LGBM exhibits also a builtin method to process categorical features. Categories are sorted with respect to the objective. Then, they use \cite{James1992} to find an optimal split according to accumulated gradients (over Hessians) in reasonable time. See below for more details.

\subsubsection{CatBoost}
Catboost \cite{DBLP:journals/corr/DorogushGGKPV17}, \cite{DBLP:journals/corr/abs-1810-11363} belongs to the class of gradient boosting algorithms and is specially suitable to input spaces that contain categorical features. Two innovations are implemented:
\begin{itemize}
    \item ordered boosting
    \item algorithm that processes categorical features
\end{itemize}
The idea of ordered boosting is to compute gradients or residuals for a given sample of the training set based on a set of samples that does not include this particular training sample. Otherwise, the gradient appears to be biased. In practice, CatBoost establishes several permutations of the training set that are used for diverse training iterations. Residuals are computed based on these permutations, and models as well.

For processing categorical features, they take inspiration from target statistics (this converts categorical features into numerical values based on target averages, see below for more details). They proved that this technique is biased and came up with a solution based on ordered boosting (the solution is called ordered TS). Categorical features can then have different values according to the training iteration.

CatBoost is also innovative in a software view in the sense, it is said to be very efficient for training and inference both in CPU and GPU.
\subsection{Builtin methods}
\subsubsection{LGBM}
LGBMs have a specific way of dealing with categorical variables. At each split, categorical variables are sorted according to the training objective instead of categorical values (accumulated sum of gradient over accumulated sum of hessians for each categorical feature). The optimal split is made using \cite{doi:10.1080/01621459.1958.10501479}.

\subsubsection{CatBoost}
The builtin method is derived from the CatBoost method presented below in Section \ref{catboost_pres}.
\subsubsection{XGBoost}
There is no specific way to handle categorical variables using XGBoost. It is naturally achieved by splitting categories according to categorical values.

\subsection{Encoding categorical variables using target statistics}
\subsubsection{General equation of target statistics}
We assume that $X$ is a categorical variable which takes $I$ different categories and that Y is a binary target in a supervised learning context.
We use the following notations: 
\begin{itemize}
    \item  $n_{i Y}$ is the number of times $X=i$ when $Y=1$
    \item $n_{i}$ is the number of times $X=i$
    \item $n_{Y}$ is the number of times $Y=1$
    \item $n_{T R}$ is the number of samples in the training set
\end{itemize}
Then, the prior probability of fraud (without taking categories into consideration) is given by:
\begin{equation}
    p_{prior} =\frac{n_{Y}}{n_{T R}}.
\end{equation}

and the probability of fraud  given categories is given by:
\begin{equation}
p_{fraud,i}=\frac{n_{i Y}}{n_{i}}.
\end{equation}

Then, the equation that allows to compute $S_{i}$ the encoded value for the category $i$ can be computed as a weighted sum of $p_{prior}$ and $p_{fraud,i}$ (taken from \cite{10.1145/507533.507538}): 
\begin{multline}
    S_{i}= \lambda\left(n_{i}\right) p_{fraud,i} + \left(1-\lambda\left(n_{i}\right)\right) p_{prior}=\lambda\left(n_{i}\right) \frac{n_{i Y}}{n_{i}} \\ +   \left(1-\lambda\left(n_{i}\right)\right) \frac{n_{Y}}{n_{T R}}.
\end{multline}

Various encoding methods are derived from this equation and they mainly differ in two aspects:
\begin{itemize}
    \item the way $\lambda\left(n_{i}\right)$ is computed
    \item the way $p_{fraud,i}$ is computed 
\end{itemize}

\subsubsection{Target encoder}
$\lambda\left(n_{i}\right)$ is computed this way (according to \cite{10.1145/507533.507538}):
\begin{equation}
    \lambda(n_{i})=\frac{1}{1+e^{- \frac{(n_{i}-k)}{f}}},
\end{equation}

where $k$ and $f$ are two tunable parameters. $k$ is half the minimal sample size for which we trust the estimate of $p_{fraud,i}$ and $f$ is a smoothing parameter. In the target encoder package, the two default values are 1 and 1.
The idea of this formula is simply to trust more $p_{fraud,i}$ when $n_{i}$ is high with respect to the k hyper-parameter.
In the following, the Target encoder is also called Barecca encoder from the name of one of its authors.

\subsubsection{M-estimate}
$\lambda\left(n_{i}\right)$ is computed this way (according to \cite{10.1145/507533.507538}):

\begin{equation}
    \lambda(n_{i})=\frac{n_{i}}{m+n_{i}},
\end{equation}

which gives the following formula for $S_{i}$:
\begin{equation}
S_{i}=\frac{n_{i}\times p_{fraud,i}+p_{prior} m}{n_{i}+m} = \frac{n_{i Y}+p_{prior} m}{n_{i}+m},
\end{equation}
where $m$ is a tunable parameter.

The rationale of M-estimate is the same as for the target encoder. When $n_{i}>>m$ then $\lambda(n_{i})$ tends to 1 and  the estimate of $p_{fraud,i}$ becomes more important.

\subsubsection{CatBoost encoding}
\label{catboost_pres}
For CatBoost, several ideas are mixed. First, the basic formula is given by M-estimate:
\begin{equation}
   S_{i}= \frac{n_{i Y}+p_{prior} m}{n_{i}+m}.
\end{equation}

CatBoost developers identified that using the sample $n$ to compute $S_{i,n}$ leads to target shift. To avoid using the sample $n$, they use permutations of training sets and take systematically all the relevant elements of the permuted dataset before the new position of sample $n$ to compute $n_{i}$ or $n_{i Y}$.

\subsubsection{Pozzolo}
We take inspiration from \cite{Pozzolo2015AdaptiveML} which designs a categorical encoding method specially designed for the problem of credit card fraud detection.
$\lambda\left(n_{i}\right)$ is computed this way:

Let $\alpha_{n_{i}}$ defined as:
\begin{equation}
    \alpha_{n_{i}}= \frac{n_{i}}{n_{T R}}.
\end{equation}

Then, we define two ways of computing $\lambda\left(n_{i}\right)$:
\begin{equation}
    \lambda1(n_{i})= \frac{\alpha_{n_{i}}- \min_{i\in I}(\alpha_{n_{i}})}{\max_{i\in I}(\alpha_{n_{i}})-\min_{i\in I}(\alpha_{n_{i}})},
\end{equation}
and
\begin{equation}
    \lambda2(n_{i})= \frac{log(\alpha_{n_{i}}- \min_{i\in I}(\alpha_{n_{i}}))}{log(\max_{i\in I}(\alpha_{n_{i}})-\min_{i\in I}(\alpha_{n_{i}}))}.
\end{equation}

The log computation is made such that the coefficients take into account the fact that the values of categorical variables are not uniformly distributed over the categories, which is often the case for categorical variables in fraud detection data.

For instance, let's say that X has three categories 'A', 'B' and 'C'. 'A' is called 99 \% of the samples, and 'B' and 'C' 0.5 \% of the samples. Then, using $\lambda1(n_{i})$ will yield very large values for 'A' and very small ones for 'B' and 'C'. The log computation used for $\lambda2(n_{i})$ makes these differences smaller.

\subsubsection{James-Stein}

This encoder directly comes from the Stein's paradox \cite{James1992} (the methods based on target statistics have a connection with the paradox) which tells that combined estimators are more accurate on average with respect to a classical Monte-Carlo estimator.

The classical James Stein approach gives the following formula for $1-\lambda\left(n_{i}\right)$:

\begin{equation}
    1-\lambda\left(n_{i}\right)=\frac{(k-3) var(p_{fraud,i})}{\sum\left(x_{i}-p_{prior}\right)^{2}},
\end{equation}
where k is the number of groups. 

$(k-3)$ has been replaced by $k-1$ and we found the following formula:
\begin{multline}
    1-\lambda\left(n_{i}\right)=\frac{(k-1) var(p_{fraud,i})}{\sum\left(x_{i}-p_{prior}\right)^{2}}= \\ \frac{ var(p_{fraud,i})}{\frac{\sum\left(x_{i}-p_{prior}\right)^{2}}{k-1}}=\frac{ var(p_{fraud,i})}{var(p_{fraud,i})+var(p_{prior})}.
\end{multline}

Then, $\lambda\left(n_{i}\right)$ is computed this way:
\begin{equation}
\lambda\left(n_{i}\right) = \frac{var(p_{prior})}{var(p_{fraud,i})+var(p_{prior})},
\end{equation}

where $\lambda$ depends on $n_{i}$ because $var(p_{fraud,i})$ depends on $n_{i}$. Indeed, in practice, squared standard errors are used to compute these variances.
Here, in order to compute $\lambda$, we take into account variances instead of only the number of samples taking $i$ as a category. 

\subsection{Encoding categorical variables using weight-of-evidence}
This encoding method does not depend on the equation of target statistics and comes from the credit scoring world where log ratio of percentages of bad customers and percentages of good customers \cite{articlewoe} are computed.

Let's define first weight of evidence as a log ratio between percentages of non-events (non-fraud) and events (fraud): 
\begin{equation}
woe = \log(\frac{\% of nonfrauds}{\% of frauds}).
\end{equation}

We apply this definition for each category i and $S_{i}$ is computed: 
\begin{equation}
S_{i} = \log(\frac{\frac{n_{i \bar{Y}}}{n_{\bar{Y}}}}{\frac{n_{i Y}}{n_{Y}}}).
\end{equation}

In a nutshell, this encoding method groups categories whose ratio between percentages of non-frauds and frauds is similar.

In practice, the real implemented formula is:
\begin{equation}
    S_{i} = \log(\frac{\frac{n_{i \bar{Y}}+ \gamma}{n_{\bar{Y}}+2\gamma}}{\frac{n_{i Y}+\gamma}{n_{Y}+2\gamma}}),
\end{equation}

where $\gamma$ is a regularization parameter.

\subsection{Toy example}
\begin{figure}[!h]
    \centering
    \includegraphics[width=0.45\textwidth]{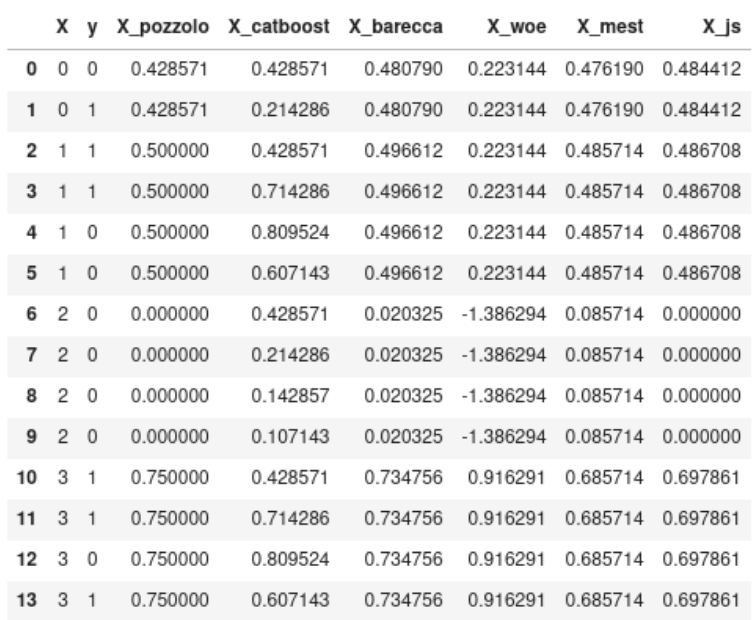}
    \caption{Toy example for categorical feature computation}
    \label{ex_cat}
\end{figure}

Figure \ref{ex_cat} shows a toy example of encodings of categorical variables. The columns X and y represent the toy inputs and outputs respectively, and the other columns stand for encodings of Pozzolo, CatBoost, Barecca (Target encoder), Weight of evidence, M-estimate and James-Stein.

\section{Experiments}

\subsection{Settings}

\subsubsection{Datasets}

We use real data from a major French bank.
More precisely, we use a subset of 10 millions of transactions using raw and derivative features. The raw features consist of 8 categorical features and 2 numerical features. The derivative features consist of a small subset of 20 numerical features among more than 200 computed using count, mean and diff aggregations.

\subsubsection{Methodology}
The datasets are splitted into training and validation sets according to a $\frac{2}{3}/ \frac{1}{3}$ policy.
For each gradient boosting model (LGBM, XGBoost, CatBoost), we test the encoding methods that are mentioned above. 
Furthermore, hyperparameters of these gradient boosting models have been optimized using the following approach: 

 The optimization happened using 3-fold cross validation on the training set (the optimization criterion was the PR AUC) using the Optuna package \cite{optuna_2019} and the Tree-Parzen-Estimator as the sampling strategy \cite{NIPS2011_86e8f7ab}. For this optimization, we did not use categorical encoding. 

For each method, we apply encoding computed using training data only. The methods are evaluated using various metrics:
\begin{itemize}
    \item Area under Precision-recall curve (PR AUC)
    \item Precision
    \item Recall
    \item F1
\end{itemize}
As ranks between methods are not stable over the seeds given as input to the boosting models, we average each setting over 10 seeds. We did not use the accuracy metric since it does not make much sense in the context of fraud detection database which is highly imbalanced.

The gradient boosting hyperparameters have been found only for the case of the absence of encoding. Then, we willingly favor the absence of encoding compared with the use of categorical encoding.

\subsection{Results}
We display the results metrics by using a table gathering average results of each method.
More precisely, we only display percentage of variations with respect to the absence of encoding. We willingly omit raw values for confidentiality purposes.
\subsubsection{LGBM}
\begin{table}[h!]
    \centering
    \begin{tabular}{|p{1.5cm}|c|c|c|c|}
    \hline
         &  PR AUC & Precision & Recall & F1 \\
    \hline
    Pozzolo    & +9\% & -1\% & +59\% & +53\% \\
     \hline
    M-estimate    & +6\% & -7\% & +61\% & +53\% \\
     \hline
    Builtin    & \textbf{+11\%} & +0\% & +36\% & +32\% \\
     \hline
    James Stein    & +5\% & -4\% & \textbf{+64\%} &  \textbf{+56\%} \\
     \hline
    Barecca    & +9\% & +0\% & +57\% & +50\% \\
     \hline
    Weight of evidence    & +8\% & +1\% & +56\% & +50\% \\
     \hline
    CatBoost    & +5\% & \textbf{+3\%} & +41\% & +37\% \\
     \hline
    \end{tabular}
    \caption{Results for LGBM}
    \label{lgbm_res}
\end{table}

    For LGBM, we establish several observations:
    \begin{itemize}
        \item We observe that all the encoding methods improve PR AUC, RECALL and F1.
        \item We observe that only CatBoost and Weight of Evidence improve the Precision metric.
        \item The builtin method gives very interesting results without implementation effort due to the encoding.
    \end{itemize}
\subsubsection{CatBoost}
\begin{table}[h!]
    \centering
    \begin{tabular}{|p{1.5cm}|c|c|c|c|}
    \hline
         &  PR AUC & Precision & Recall & F1 \\
    \hline
     Barecca    & +27\% & -8\% & +106\% & +96\% \\
     \hline
     Builtin    & \textbf{+34\%} & -9\% & \textbf{+133\%} & \textbf{+119\%} \\
     \hline
     Pozzolo & +26\% & -8\% & +91\% & +82\% \\
     \hline
     James-Stein & +24\% & -11\% & +114\% & +102\% \\
     \hline
     Weight of evidence & +22\% & \textbf{-4\%} & +102\% & +92 \% \\
     \hline
     M-estimate & +22\% & -11\% & +95\% &+86\% \\
     \hline
     
    \end{tabular}
    \caption{Results for CatBoost}
    \label{cb_res}
\end{table}
        For CatBoost, we establish several observations:
    \begin{itemize}
        \item Precision is slightly deteriorated by all the methods though all the other metrics are improved.
        \item The builtin Catboost method is the best one according to all the metrics (with the exception of Precision) and should be chosen in the future since it can be applied in an effortless way.

     \end{itemize}
\subsubsection{XGBoost}
        For XGBoost, we establish several observations:
    \begin{itemize}
        \item The CatBoost encoder method is the only one to improve all the metrics, precision included.
        \item Another interesting method is the target encoder (Barecca).
     \end{itemize}
    To summarize, all these results show us on the one hand that gradient boosting algorithms applied on real fraud detection problems benefit from categorical encoding. On the other hand, we notice that the CatBoost encoder seems to be a very good categorical encoding method for all the tested gradient boosting models.
\begin{table}[h!]
    \centering
    \begin{tabular}{|p{1.5cm}|c|c|c|c|}
    \hline
         &  PR AUC & Precision & Recall & F1 \\
    \hline
     M-estimate    & +5\% & -5\% & +49\% & +44\% \\
     \hline
     CatBoost    & \textbf{+6\%} & \textbf{+4\%} & +36\% & +29\% \\
     \hline
     Pozzolo    & +4\% & -5\% & +40\% & +36\% \\
     \hline
     James Stein    & 0\% & -9\% & \textbf{+53\%}& \textbf{+47\%} \\
     \hline
    Barecca    & +6\% & -4\% & +46\% & +41\%\\
    \hline
    Weight of evidence   & +1\% & -9\% & +36\% & +32\% \\
     \hline
    \end{tabular}
    \caption{Results for XGBoost}
    \label{xgb_res}
\end{table}
\section{Conclusion}

We have tested several categorical encoding approaches for different gradient boosting methods. First, we have observed that it is always interesting and beneficial to handle categorical variables in a specific way as metrics are all improved with the exception of precision in some cases. Then, we back up the claim that gradient boosting algorithms also benefit from the encoding of categorical variables. Second, we remind the reader of the fact that these results are only valid for the specific database, for the specific categorical encoding parameters and for the gradient-boosting methods along with its parameters.

Nevertheless, we observe that for LGBM, Catboost and Weight-of-evidence methods yield best performances. The builtin method is also interesting and can be applied effortlessly. For CatBoost, the builtin method is by far the best one. For XGBoost, CatBoost and Barecca encoding methods seem to be the best ones.
Consequently, we claim that the CatBoost method seems to be the most promising one for encoding categorical variables in our real fraud detection problem.

For future work, we wish to consolidate the results using different datasets and categorical encoding hyper-parameter optimization.

\newpage

\printbibliography

\end{document}